%% file: iclr2024_conference.tex
\documentclass{article} 
\usepackage{iclr2024_conference,times}
\usepackage{amsmath,graphicx,caption}

\input{math_commands.tex}

\usepackage{hyperref}
\usepackage{url}

\title{Constricting Normal Latent Space for Anomaly Detection with Normal-only Training Data}

\author{Marcella Astrid$^{1,2,3}$ \qquad Muhammad Zaigham Zaheer$^{4}$ \qquad Seung-Ik Lee$^{1,2 *}$  \\
$^{1}$Department of Artificial Intelligence, \\ University of Science and Technology, South Korea\\
$^{2}$Field Robotics Research Section, \\Electronics and Telecommunications Research Institute, South Korea \\
$^{3}$Interdisciplinary Centre for Security, Reliability and Trust, \\ University of Luxembourg, Luxembourg \\
$^{4}$Department of Computer Vision, \\ Mohamed Bin Zayed University of Artificial Intelligence, United Arab Emirates\\
\texttt{marcella.astrid@uni.lu} \\
\texttt{zaigham.zaheer@mbzuai.ac.ae} \\
\texttt{the\_silee@etri.re.kr} \qquad
$^{*}$corresponding author
}

\iclrfinalcopy 
\begin{document}

\maketitle

\begin{abstract}
In order to devise an anomaly detection model using only normal training data, an autoencoder (AE) is typically trained to reconstruct the data. As a result, the AE can extract normal representations in its latent space. During test time, since AE is not trained using real anomalies, it is expected to poorly reconstruct the anomalous data. However, several researchers have observed that it is not the case. In this work, we propose to limit the reconstruction capability of AE by introducing a novel latent constriction loss, which is added to the existing reconstruction loss. By using our method, no extra computational cost is added to the AE during test time. Evaluations using three video anomaly detection benchmark datasets, i.e., Ped2, Avenue, and ShanghaiTech, demonstrate the effectiveness of our method in limiting the reconstruction capability of AE, which leads to a better anomaly detection model.
\end{abstract}

%
%

\vspace{-2mm}
\section{Introduction}
\label{sec:intro}

Anomaly detection is a challenging problem due to the rarity of anomalous event occurrences. Therefore, it is typically approached as one class classification (OCC) problem where data instances of only one class, i.e., normal class, are used in the training \citep{hasan2016learning,zaheer2020old,zhao2017spatio,luo2017remembering,georgescu2021anomaly,park2021anomaly}. In this setting, an autoencoder (AE) is generally used to encode the normalcy representations in its latent space \citep{hasan2016learning,zhao2017spatio,luo2017remembering,park2020learning,gong2019memorizing}. Specifically, the AE is trained to reconstruct the normal data given in the training set. Since AE is not trained using real anomalies, it is expected to not reconstruct anomalies during test time. 
Therefore, the normal and anomalous data should be distinguishable by the difference in their reconstruction quality.
However, as observed by several researchers \citep{astrid2021learning,astrid2021synthetic,gong2019memorizing}, AE can often `generalize' so well so it can start reconstruct any input, including the anomalous data. This property reduces the capability of an AE to discriminate between normal and anomalous data. It occurs because there is no mechanism constricting the latent space, which enables the latent space to grow too large. Consequently, the AE adapts and learns to reconstruct from this large latent space, which can cover the anomalous data as well.

To prevent this from happening, several researchers utilize memory mechanism to limit the latent space \citep{gong2019memorizing,park2020learning}. 
The AE is bound to reconstruct data using latent space within the span of memory vectors. 
However, as mentioned in \citep{gong2019memorizing}, this requires additional computational costs to read the memory during test time. Rather than limiting the latent space using memory vectors, in this work, we propose to approach the problem directly by introducing a new training loss that constricts the latent space. 
In this way, AE learns to produce reconstructions using only the constricted latent space.
Moreover, since we only add an additional loss during training time, there is no extra computational cost during test time.

In summary, our contributions are as follows: 1) We propose a novel constriction loss to limit the reconstruction capability of AE without any additional component added to the architecture; 2) We introduce and explore two types of the proposed constriction loss; 3) We evaluate the effectiveness of our method on three video anomaly benchmark datasets, i.e., Ped2 \citep{li2013anomaly}, Avenue \citep{lu2013abnormal}, and ShanghaiTech \citep{luo2017revisit}.


\vspace{-2mm}
\section{Related Works}
\vspace{-3mm}


To limit the reconstruction capability of AE, in addition to the aforementioned memory-based methods \citep{park2020learning,gong2019memorizing}, several existing studies have utilized pseudo anomalies \citep{astrid2021learning,astrid2021synthetic}.
However, such methods require prior knowledge to generate the pseudo anomaly. On the other hand, our method does not require it. We also acknowledge other methods which are not reconstruction-based methods. For example, binary classifier \citep{zaheer2020old,pourreza2021g2d,georgescu2021anomaly} and prediction task \citep{liu2018future,georgescu2021anomaly,dong2020dual}. Additionally, there are methods that utilize real anomalous data \citep{zaheer2023clustering,munawar2017limiting,sultani2018real}. Our method, on the other hand, uses reconstruction-based method and only normal data for the training.


\begin{figure}
\centering
\begin{minipage}{.47\textwidth}
  \centering
  \includegraphics[width=.8\linewidth]{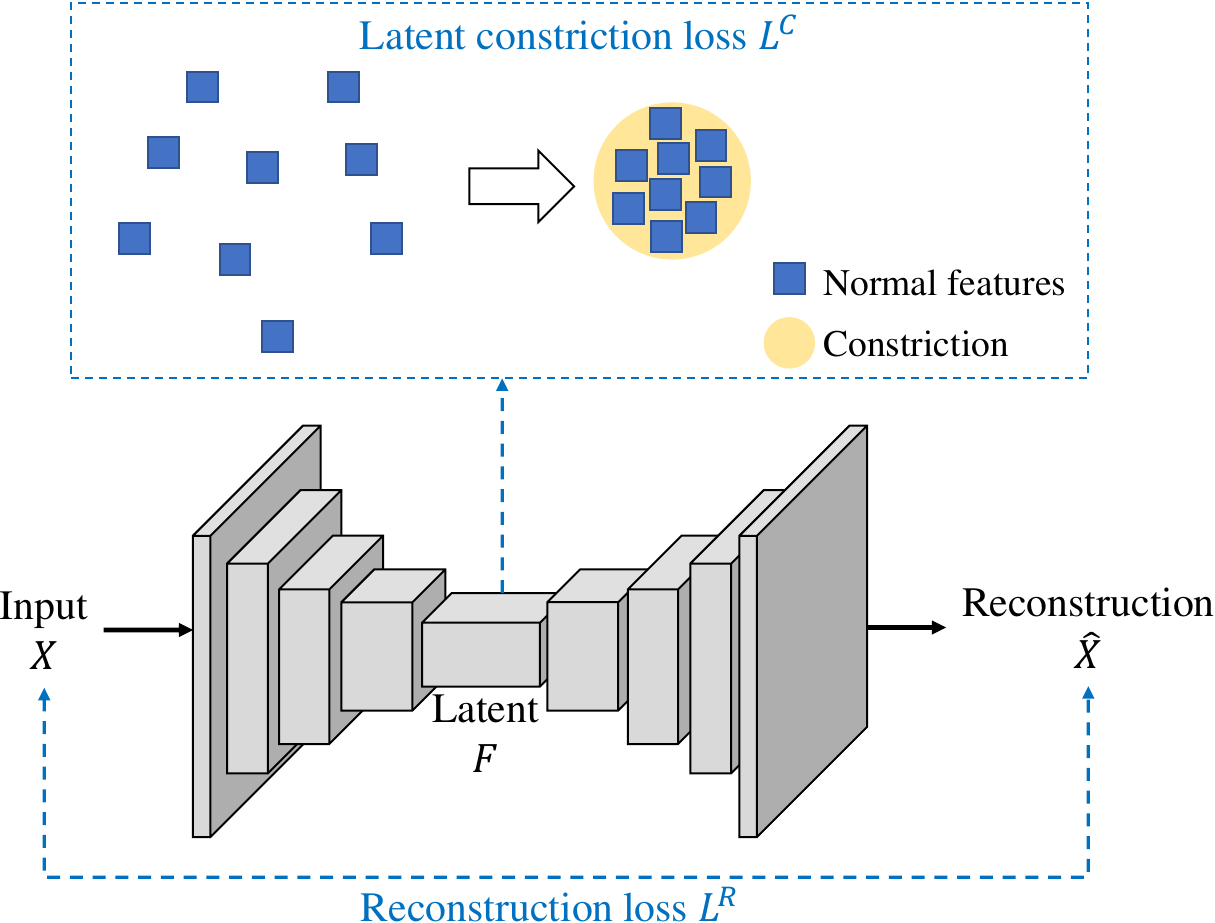}
  \captionof{figure}{Overall configuration of our proposed method which consists of an autoencoder (AE) trained using reconstruction loss and our proposed latent constriction loss. The latent constriction loss restrains the normal features into smaller space in order to limit the reconstruction capability of the AE.}
  \label{fig:overall}
\end{minipage}%
\hspace{0.06\textwidth}%
\begin{minipage}{.47\textwidth}
  \centering
  \includegraphics[width=\linewidth]{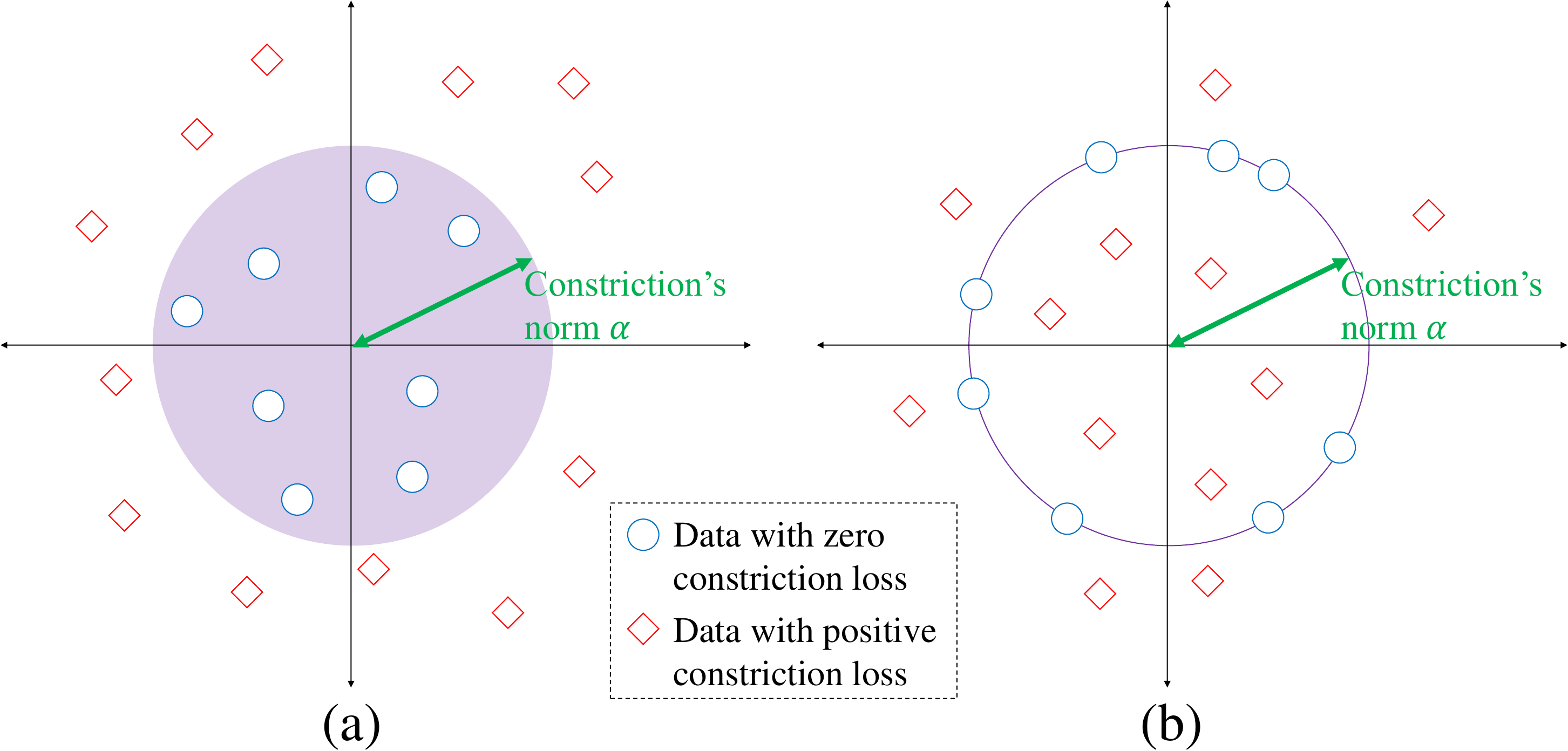}
  \captionof{figure}{Two types of the proposed latent constriction losses: (a) constricting inside the norm sphere, (b) constricting on the surface of the norm sphere.}
  \label{fig:constriction}
\end{minipage}
\vspace{-6mm}
\end{figure}

\section{Methodology}
\vspace{-2mm}

In this section, we discuss each component of our approach as seen in Figure \ref{fig:overall}, including conventional AE trained using reconstruction loss (Section \ref{subsec:conventionalAE}), the novel constricting loss (Section \ref{subsec:constriction}), and the inference technique (Section \ref{subsec:inference}).

\subsection{Training AE to represent normal data}
\label{subsec:conventionalAE}
\vspace{-2mm}

In OCC setting, an AE is typically used to create representation of normal data. The AE takes input $X$ of size $T \times C \times H \times W$ and outputs its reconstruction $\hat{X}$ of the same size, where $T$, $C$, $H$, $W$ are respectively the number of frames, number of channels, height, and width of the input. 
An AE consists of an encoder $\mathcal{E}$ and a decoder $\mathcal{D}$. $\mathcal{E}$ encodes $X$ into a latent feature $F$ of size $T' \times C' \times H' \times W'$. $\mathcal{D}$ then decodes $F$ into $\hat{X}$. To learn the normal representation in the latent space, conventionally, the AE is trained to reconstruct normal data available in the training data as:
\begin{equation}
    L^R= \frac{1}{T \times C \times H \times W}  \left \| \hat{X} - X  \right \|_{F}^{2} \text{,}
\label{eq:aereconloss}
\end{equation}
where $\left \| \cdot \right \|_{F}$ is Frobenius norm.

\subsection{Constricting the normal data latent space}
\label{subsec:constriction}
\vspace{-2mm}

Since the AE is trained using only normal data, the latent features ideally consists of only normal representations. However, as nothing restricting the latent features, the latent space can grow too large, which may include lot of anomalous data features. This can lead to the problem where AE can reconstruct anomalous data as well as the normal data. Therefore, we propose to limit the latent space by adding a new constriction loss. 

The overall loss in our method combines the reconstruction loss $L^R$ in \eqref{eq:aereconloss} and our proposed constriction loss $L^C$ as:
\begin{equation}
    L = L^R + \lambda L^C \text{,}
\label{eq:totalloss}
\end{equation}
where $\lambda$ is the loss weighting hyperparameter. 

$L^C$ specifically constricts the latent features $F_{j,i} \in \mathbb{R}^{T' \times C'}$ taken from $F$. 
We combine the channel $C'$ and time $T'$ dimension to incorporate both appearance and time information inside the latent features that we optimize. In this work, we propose two types of constriction losses using norm sphere which we discuss in details next.


\subsubsection{Constricting inside the sphere}
\vspace{-2mm}
For the first type of constriction loss, we constrict the normal data to be inside a norm sphere. Therefore, $L^C$ is defined as:
\begin{equation}
    L^C = \frac{1}{H' \times W'} \sum_{j=1}^{H'} \sum_{i=1}^{W'} \max(0, \left \| F_{j,i}  \right \|_{2} - \alpha) \text{,}
\label{eq:insidesphere}
\end{equation}
where $\left \| \cdot  \right \|_{2}$ is L2-norm and $\alpha$ is a hyperparameter to control the size of the sphere.  If $\left \| F_{j,i}  \right \|_{2} \leq \alpha$, the $F_{j,i}$ is already inside the norm sphere, hence the constriction loss is zero. Otherwise, the loss is positive. Illustration of the loss can be seen in Figure \ref{fig:constriction}(a).

\subsubsection{Constricting on the surface of the sphere}
\vspace{-2mm}
Another way to limit the features is by constricting on the surface of a norm sphere as: 
\begin{equation}
    L^C = \frac{1}{H' \times W'} \sum_{j=1}^{H'} \sum_{i=1}^{W'} \left| \left \| F_{j,i}  \right \|_{2} - \alpha \right| \text{,}
\label{eq:onsphere}
\end{equation}
where $\left| \cdot \right|$ is absolute operation. $L^C$ is zero only if $\left \| F_{j,i}  \right \|_{2} = \alpha$. Outside of the surface, whether by having smaller or larger norm, the loss will be positive. Illustration of the loss can be seen in Figure \ref{fig:constriction}(b).

\subsection{Inference}
\label{subsec:inference}
\vspace{-2mm}

During test time, we follow the inference mechanism used in several reconstruction-based methods \citep{park2020learning,astrid2021learning,astrid2021synthetic} by utilizing PSNR value $\mathcal{P}_t$ of $t$-th input frame $I_t$ and its reconstruction $\hat{I}_t$. $\mathcal{P}_t$ is then min-max normalized across each test video to produce a normalcy score $\mathcal{N}_t$ of range $[0, 1]$. Finally, the anomaly score the frame is calculated as: 
\begin{equation}
    \mathcal{A}_t = 1-\mathcal{N}_t \textbf{.}
\end{equation}
In this way, a higher value of $\mathcal{A}_t$ indicates a higher level of abnormality found in $I_t$.

\vspace{-2mm}
\section{Experiments}
\vspace{-2mm}
\subsection{Datasets}

We evaluate our method on three video anomaly benchmark datasets, i.e., Ped2 \citep{li2013anomaly}, Avenue \citep{lu2013abnormal}, and ShanghaiTech \citep{luo2017revisit}. The training set of each of the dataset consists only of normal frames, whereas each video in the test set consists of one or more anomalous portions. 

\noindent\textbf{Ped2.} It contains 16 training and 12 test videos. Normal behavior is defined as walking pedestrians. Whereas abnormal behaviors include riding bicycles, skateboards, and vehicles.

\noindent\textbf{Avenue.} It consists of 16 training and 21 test videos, where walking people are considered normal. Whereas, examples of anomalous behaviors are throwing bags and running.

\noindent\textbf{ShanghaiTech.} It is by far the largest one-class anomaly detection dataset consisting of 330 training and 107 test videos recorded at 13 different locations with various camera angles and lighting conditions. In total, there are 130 anomalous events in the test set including running, riding bicycle, and fighting.

\subsection{Experimental setup}

\noindent\textbf{Evaluation metric.} We utilize a widely used frame-level area under the ROC curve (AUC) to evaluate our method. The ROC curve is computed for all videos in each test set, i.e., one ROC curve for each dataset.

\noindent\textbf{Implementation details.} As the baseline AE, we use the baseline in \citep{astrid2021learning} and train it using only reconstruction loss (i.e., $\lambda = 0$ in \eqref{eq:totalloss}). In the AE trained using our constriction loss, we remove the last LeakyReLU of the encoder to let the latent space covers the negative and positive space equally before constricting it with our proposed loss. The input and output of each model have size of $16 \times 1 \times 256 \times 256$. The training configurations, such as mini batch size, learning rate, and optimizer, follow \citep{astrid2021learning}. During inference (Section \ref{subsec:inference}), we compute anomaly score using the 9th frame. By default, for all experiments, we set the loss weighting hyperparameter to $\lambda=0.0001$. As for constricting inside the sphere, we set $\alpha=0.1$ for Ped2 and $\alpha=1$ for the other datasets. Whereas for constricting on the surface of the sphere, we set $\alpha=10$ for Ped2 and $\alpha=100$ for the other datasets. We repeat our experiments on both the baseline and our method five times and select the maximum performance.

\subsection{Comparisons with the baseline}
\vspace{-2mm}

\begin{figure*}[htb]
  \centering
  \includegraphics[width=\linewidth]{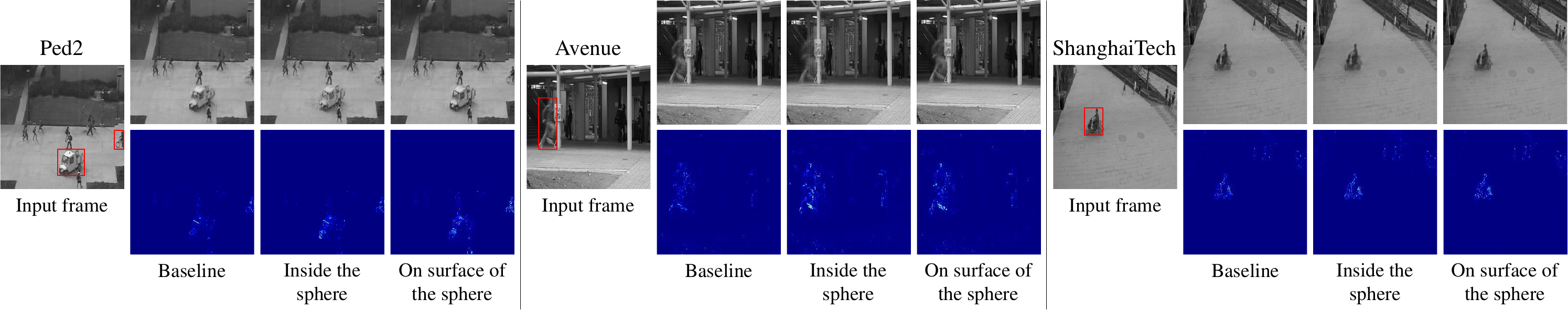}
  \vspace{-4mm}
  \caption{Qualitative comparisons of the baseline and our method in test samples from each dataset. The top and bottom rows are the outputs of the AE and the respective reconstruction error heatmaps. Reconstruction error heatmaps are computed using reconstruction error in the frame followed by min-max normalization in a frame. Red boxes mark the anomalous regions. Our method successfully distorts the anomalous regions better than the baseline.}
  \vspace{-3mm}
  \label{fig:qualitative}
\end{figure*}

To find out the effectiveness of our method in limiting reconstruction capability of AE, we compare our method with the baseline AE. The AUC comparisons can be seen in the last three rows of Table \ref{tab:sota}. Each of our method, using either of the constriction methods, successfully outperforms the baseline which demonstrates the effectiveness of our approach.
Moreover, as seen qualitatively in Figure \ref{fig:qualitative}, our method successfully limits the reconstruction capability of AE which consequently produces higher reconstruction error in the anomalous regions compared to the baseline.

\begin{table}[]
\centering
\resizebox{0.6\linewidth}{!}{
\small
\begin{tabular}{|l|ccc|}
\hline
Methods                 & Ped2   & Ave  & Sh      \\ \hline \hline
  AE-Conv2D  \citep{hasan2016learning}          & 90.0\%  & 70.2\%  & 60.85\% \\
  AE-Conv3D  \citep{zhao2017spatio}             & 91.2\%  & 71.1\%  & -       \\
  AE-ConvLSTM  \citep{luo2017remembering}       & 88.10\% & 77.00\% & -       \\
  TSC \citep{luo2017revisit}                    & 91.03\% & 80.56\% & 67.94\% \\
  StackRNN \citep{luo2017revisit}               & 92.21\% & 81.71\% & 68.00\% \\
  STEAL Net \citep{astrid2021synthetic}      & \textbf{98.4\%}  & \textbf{87.1\%} & \underline{73.7\%} \\
  LNTRA - Patch \citep{astrid2021learning}      & 94.77\%  & \underline{84.91\%} & 72.46\% \\
  LNTRA - Skip frame \citep{astrid2021learning} & \underline{96.50\%}  & 84.67\% & \textbf{75.97\%} \\
  MemAE \citep{gong2019memorizing}              & 94.1\%  & 83.3\%  & 71.2\%  \\
  MNAD-Reconstruction \citep{park2020learning}           & 90.2\%  & 82.8\%  & 69.8\%  \\
  \hline
  Baseline             & 92.49\%  & 81.47\% & 71.28\% \\
  Ours-Inside Sphere       & 94.66\% & 82.93\% & 71.35\% \\
  Ours-On Sphere Surface        & 94.05\% & 82.14\% & 71.39\%  \\
   \hline

\end{tabular}
}
\vspace{-2mm}
\caption{AUC performance comparison of our approach with several existing reconstruction-based SOTA methods on Ped2, Avenue (Ave), and ShanghaiTech (Sh). 
Best and second best performances are highlighted as bold and underlined. 
}
\vspace{-2mm}
\label{tab:sota}
\end{table}

 \begin{figure}[htb]
  \centering
  \includegraphics[width=\linewidth]{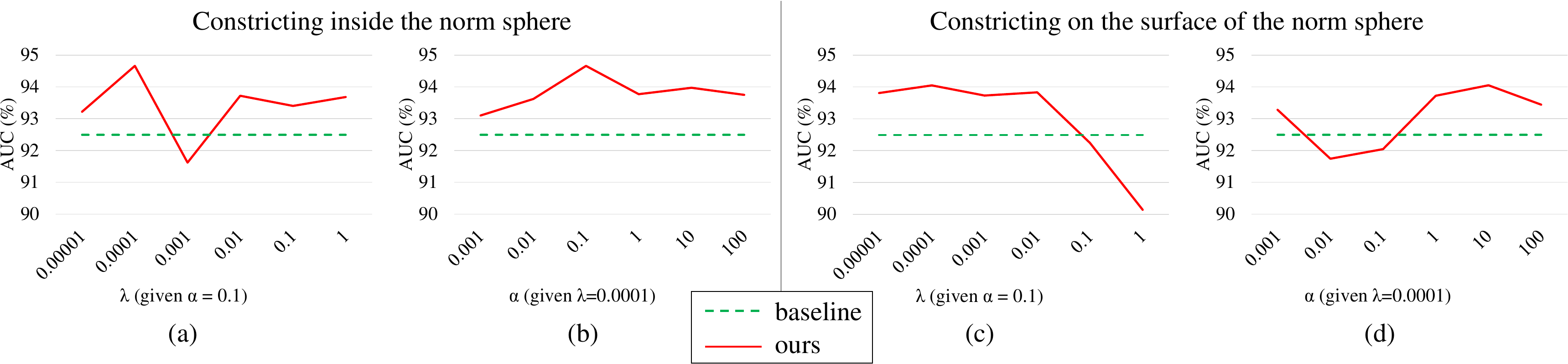}
  \vspace{-4mm}
  \caption{Evaluation of the hyperparameters used in our method on Ped2 dataset in constricting (a)-(b) inside the norm sphere and (c)-(d) on the surface of the norm sphere. Different loss weighting $\lambda$ (\eqref{eq:totalloss}) and constricting norm $\alpha$ (\eqref{eq:insidesphere} \& \eqref{eq:onsphere}) are used. As mostly our method (red solid line) outperforms the baseline (green dotted line), our method is robust towards different hyperparameter values.}
  \vspace{-3mm}
  \label{fig:hyperparameter}
\end{figure}

\subsection{Comparisons with other methods}
\vspace{-2mm}
For fair comparisons, we compare our approach only to reconstruction-based methods (i.e., methods that use reconstruction as the sole decision factor of anomaly score) in Table \ref{tab:sota}.

As seen, our method is on par with memory-based networks, i.e., MemAE and MNAD-Reconstruction. With similar performance, our method does not required any additional computation cost on top of the baseline AE during test time, while memory based networks require memory reading operation \citep{gong2019memorizing}. This also demonstrates that directly limiting the latent space using constriction loss performs similarly to limiting using memory.

However, compared to method utilizing pseudo anomalies, such as STEAL Net \citep{astrid2021synthetic} and LNTRA \citep{astrid2021learning}, our method is inferior. In spite of that, our method does not require any prior knowledge while still achieving competitive performance. On the other hand, STEAL Net \citep{astrid2021synthetic} and LNTRA \citep{astrid2021learning} require prior knowledge on what anomalies are, such as anomalous speed and appearance.

\subsection{Hyperparameters evaluation}

In our proposed method, we introduce two new hyperparameters, i.e., loss weighting $\lambda$ and constricting norm $\alpha$. Figure \ref{fig:hyperparameter} shows the results of our experiments using different values of hyperparameters. To limit the span of experiments, the experiments are conducted in Ped2 only. In most values, our method is robust and shows better performances compared to the baseline. 
However, constricting on the surface generally does not work with smaller $\alpha$ values because it is already very constricting compared to constricting inside the sphere. Using a smaller $\alpha$ can lead to too much constrictions, which can also hinder the AE to reconstruct any data, including the normal data itself. 

\section{Conclusion}
In this paper, we propose a novel constriction loss to limit the reconstruction capability of AE in anomaly detection. The loss is applied to the latent space between the encoder and the decoder of AE. We introduce two types of constriction losses, i.e., constricting the latent inside norm sphere and on the surface of norm sphere. The evaluations on Ped2, Avenue, and ShanghaiTech datasets demonstrate the effectiveness of our method in improving the capability of the AE in distorting the anomalous inputs. Moreover, compared to memory-based networks \citep{gong2019memorizing,park2020learning} that require the computational burden of reading memory, our method does not require any additional computation cost during test time. 

\section*{Acknowledgement}
This work was supported by Institute of Information \& communications Technology Planning \& Evaluation (IITP) grant funded by the Korea government(MSIT) (No. 2022-0-00951, Development of Uncertainty-Aware Agents Learning by Asking Questions)

\bibliography{iclr2024_conference}
\bibliographystyle{iclr2024_conference}

\end{document}

%% file: math_commands.tex
\usepackage{amsmath,amsfonts,bm}

\def\eqref#1{equation~\ref{#1}}

\def\1{\bm{1}}

\DeclareMathAlphabet{\mathsfit}{\encodingdefault}{\sfdefault}{m}{sl}
\SetMathAlphabet{\mathsfit}{bold}{\encodingdefault}{\sfdefault}{bx}{n}

